\documentclass[10pt,twocolumn,letterpaper]{article}

\usepackage{iccv}
\usepackage{times}
\usepackage{epsfig}
\usepackage{graphicx}
\usepackage{amsmath,amsfonts,amssymb}
\usepackage[table,xcdraw]{xcolor}
\usepackage{colortbl}
\usepackage{multirow}
\usepackage{verbatim}
\usepackage{hhline}
\usepackage{arydshln}

\usepackage[export]{adjustbox}
\usepackage[caption=false,font=normalsize,labelfont=sf,textfont=sf]{subfig}

\usepackage[pagebackref=true,breaklinks=true,letterpaper=true,colorlinks,bookmarks=false]{hyperref}

\iccvfinalcopy 

\def\iccvPaperID{2} 

\ificcvfinal\fi

\begin{document}

\title{Incremental Learning Techniques for Semantic Segmentation}

\author{Umberto Michieli and Pietro Zanuttigh\\
Department of Information Engineering, University of Padova, Italy\\
{\tt\small \{umberto.michieli,zanuttigh\}@dei.unipd.it}
}

\maketitle
\ificcvfinal\thispagestyle{empty}\fi

\begin{abstract}
Deep learning architectures exhibit a critical drop of performance due to catastrophic forgetting when they are required to incrementally learn new tasks. 
Contemporary incremental learning frameworks focus  on image classification and object detection while in this work we formally introduce the incremental learning problem for semantic segmentation in which a pixel-wise labeling is considered. To tackle this task we propose to distill the knowledge of the previous model to retain the information about previously learned classes, whilst updating the current model to learn the new ones. We propose various approaches working both on the output logits and on intermediate features.  In opposition to some recent frameworks, we do not store any image from previously learned classes and only the last model is needed to preserve high accuracy on these classes. 
The experimental evaluation on the Pascal VOC2012 dataset shows the effectiveness of the proposed approaches. 
\end{abstract}


\section{Introduction and Related Work}
\label{sec:intro}

Deep neural networks are a key tool for computer vision systems. Despite their wide success on many visual recognition problems, neural networks struggle in learning new tasks whilst preserving good performance on previous ones since they suffer from catastrophic forgetting \cite{french1999catastrophic, goodfellow2013empirical, mccloskey1989catastrophic}. More precisely, the incremental learning problem is defined as the capability of machine learning architectures to continuously improve the learned model by feeding new data without losing previously learned knowledge. This has been widely studied in the context of problems like image classification and object detection \cite{castro2018end,li2018learning,rebuffi2017icarl,shmelkov2017incremental,wu2018incremental}. 
Traditional learning models  require that all the samples corresponding to old and new tasks are available during all steps of the training stage;  
a real world system, instead, should be able to update its knowledge with few training steps incorporating the new tasks while preserving unaltered the previous ones. Such a behavior is inherently present in human brain which is incremental in the sense that new tasks are continuously incorporated but the existing knowledge is  preserved.

Catastrophic forgetting represents one of the main limitations of neural networks. It has been addressed even before the rise of neural networks popularity \cite{cauwenberghs2001incremental,polikar2001learn,thrun1996learning}, but more recently it has been rediscovered and tackled in different ways.
Some methods \cite{istrate2018incremental, roy2018tree, sarwar2017incremental, xiao2014error} exploit network architectures which grow during the training process.
A different strategy consists in freezing or slowing down the learning process on some relevant parts of the network \cite{istrate2018incremental, kirkpatrick2017overcoming, li2018learning, oquab2014learning}.
Another way of retaining high performance on old tasks  is knowledge distillation. This idea was originally proposed in \cite{bucilua2006model, hinton2015distilling} and then adapted in different ways in recent studies \cite{castro2018end, furlanello2016active, li2018learning, rebuffi2017icarl, shmelkov2017incremental, wu2018incremental, zhou2019M2KD} to maintain stable the responses of the network on the old tasks whilst updating it with new training samples. However, differently from this paper, previous works focus only on object detection or image classification problems.

Some studies keep a small portion of data belonging to previous tasks and use them to preserve the accuracy on old tasks when dealing with new problems \cite{castro2018end, chaudhry2018riemannian, hou2018lifelong, lopez2017gradient, rebuffi2017icarl, tasar2018incremental}. The exemplar set to store is chosen at random or according to a relevance metric. 
In \cite{castro2018end} the classifier and the features for selecting the samples to be added in the representative memory are learned jointly and herding selection is then used.
Another method of this family is the only work considering an incremental setting for semantic segmentation \cite{tasar2018incremental}, which however focuses on a very specific setup related to satellite images and has several limitations when applied to generic semantic segmentation problems. Indeed, it considers the segmentation  as a multi-task learning problem, where a binary classification for each class replaces the multi-class labeling, and it stores some patches of previously seen images. Furthermore it assumes that training images corresponding to an incremental step only contain new classes while the capabilities on old ones are preserved by storing a subset of the old images.
For large amount of classes and wide range of applications the methodology does not scale properly.

Storing previously seen data could represent a serious limitation for certain applications where privacy issues or limited storage budgets are present.
For this reason, some recent methods \cite{shin2017continual, wu2018incremental} do not store old data but compensate this by training Generative Adversarial Networks (GANs) to generate images containing previous classes while new classes are learned. 
Some other approaches do not make use of exemplars set \cite{aljundi2018memory, kirkpatrick2017overcoming, li2018learning, shmelkov2017incremental, dhar2018learning, zhou2019M2KD}. 
In \cite{shmelkov2017incremental} an end-to-end learning framework is proposed where the representation and the classifier are learned jointly without storing any of the original training samples.
In \cite{li2018learning} previous knowledge is distilled directly from the last trained model. 
In \cite{zhou2019M2KD} the current model distills the knowledge from pruned versions of all previous model snapshots.

Even if previous studies focus on different tasks and no work has been conducted on incremental learning for dense labeling task, semantic segmentation is a key task that computer vision systems must face frequently in various applications e.g., in robotics or autonomous driving \cite{biasetton2019unsupervised,michieli2018game}.
Notice that, differently from image classification, in semantic segmentation each image contains together pixels belonging to multiple classes and the labeling is dense. In particular the pixels could represent newly added classes and previously existing ones, making the problem conceptually different from incremental learning in image classification where typically a single object is present in the image and the outcome is a unique value.
Furthermore, contrary to many existing methods, we consider the most challenging setting where images from old tasks are not stored and cannot be used to help the incremental process, which is particularly relevant for the vast majority of applications with privacy concerns or storage requirements.

In the first part of this paper we formalize the problem and we present possible settings for the incremental learning task.
Then we introduce a novel framework to perform incremental learning for semantic segmentation. 
In particular we re-frame the distillation loss concept used in other fields and we propose a novel approach where the distillation loss is applied to the intermediate features level. Furthermore, we exploited the idea of freezing the encoder part of the network to preserve the feature extraction capabilities.
 To the best of our knowledge this is the first work on incremental learning for semantic segmentation which does not retain previously seen images and that has been evaluated on standard datasets, i.e., Pascal VOC2012 \cite{pascalvoc2012}.
Experimental results demonstrate that the proposed approaches obtain high accuracy even without storing any of the previous examples thanks to the proposed distillation schemes. 


\section{Problem Formulation} 
\label{sec:problem}

The incremental learning task, when referring to semantic segmentation, can be defined as the ability of a learning system (e.g., a neural network) to learn the segmentation and the labeling of the new classes without forgetting or deteriorating too much the performance on previously learned ones. 
The performance of an incremental learning algorithm should be evaluated considering the accuracy on the new classes as well as the accuracy on the old ones. While the first should be as large as possible, meaning that the algorithm is able to learn the new classes, the second should be as close as possible to the one before the addition of the new classes, thus avoiding catastrophic forgetting. The key challenge then is how to balance between the preservation of previous segmentation and labeling knowledge and the capability of  learning the new classes. 
Additionally, the considered problem is particularly hard when no data of previous tasks can be preserved, which is the scenario of interest in the majority of the applications. 
In this work we focus on the most general incremental learning framework in which: previously seen images are not used; the new images contain examples of the unseen classes combined together with pixels belonging to the old ones;  the complexity of the approach scales well as the number of classes grows.

Let us assume that the available set of samples is  $\mathcal{D}$ and is composed of $N$ images. As usual part of the data is used for training and part for testing: we refer to the training split of $\mathcal{D}$ as $\mathcal{D}^{tr}$.
Each pixel in each image of $\mathcal{D}$ is associated to a unique class belonging to the set $\mathcal{C}= \lbrace c_0,c_1,c_2,...,c_{C-1}  \rbrace$ of $C$ possible classes.
In case a background class is present we associate it to class $c_0$ because it is considered a special class with a non-conventional behavior being present in almost all the images and having by far the largest occurrence among the elements of $\mathcal{C}$.

In the incremental learning setting we assume that we have trained our network to recognize a subset $\mathcal{S}_0 \subset \mathcal{C}$ of \textit{seen} classes using a labeled subset $\mathcal{D}_0^{tr} \subset \mathcal{D}^{tr}$, whose images contain only pixels belonging to the classes in $\mathcal{S}_0$. We then perform some incremental steps $k=1,2,...$ in which we want to recognize a new subset $\mathcal{U}_k \subset \mathcal{C}$ of \textit{unseen} classes. 
Notice that at the $k$-th incremental step the set of seen classes $\mathcal{S}_{k-1}$  is the union of all the classes previously learned and after the step we  
add the ones learned during the current step $k$: more formally, $\mathcal{S}_k=\mathcal{S}_{k-1} \cup \mathcal{U}_k$ and  $\mathcal{S}_{k-1} \cap \mathcal{U}_k = \emptyset$.
At each step  a new set of training samples is available, i.e., $\mathcal{D}_k^{tr} \subset \mathcal{D}^{tr}$, whose images contain only pixels belonging to $\mathcal{S}_{k-1} \cup \mathcal{U}_k$. The set is disjoint from previously used samples, i.e., $\left( \bigcup_{j=0,...,k-1} \mathcal{D}_j^{tr} \right) \cap \mathcal{D}_k^{tr} = \emptyset$. 
It is important to notice that, differently from image classification, images in $\mathcal{D}_k^{tr}$ could also contain classes belonging to $\mathcal{S}_{k-1}$, however their occurrence is limited since $\mathcal{D}_k^{tr}$ is restricted to consider only images containing at least one class belonging to $\mathcal{U}_k$.  Furthermore, the specific occurrence of a particular class belonging to $\mathcal{S}_{k-1}$ is highly correlated to the set of classes being added (i.e., $\mathcal{U}_k$). For example if we assume that $\mathcal{S}_{k-1} = \left\lbrace \mathit{chair}, \mathit{airplane}   \right\rbrace$ and that $\mathcal{U}_k=\lbrace \mathit{dining\ table} \rbrace$, then it is reasonable to expect that $\mathcal{D}_{k}^{tr}$ contains some images having the $\mathit{chair}$  class, that typically appears together with the $\mathit{dining\ table}$, while the class $\mathit{airplane}$ is extremely unlikely.
 
Given this scenario, there exist many different ways of sampling the set $\mathcal{U}_k \subset \mathcal{C}$ of unseen classes and of selecting the cardinality of the sets $\mathcal{U}_k$ at each step, leading to different experiments. Previous work \cite{shmelkov2017incremental} ordered the classes using the sequence provided by the creators of the dataset and analyzed the behavior of the algorithms to the addition of a single class, the addition of a batch of classes and the sequential addition of classes. Our results stick to these settings to reproduce the same scenarios.

\section{Methodology} 
\label{sec:methodology}

In this work we start by re-framing  incremental learning techniques developed for other fields in the semantic segmentation task. Then we propose some novel strategies explicitly targeted to this problem.

The proposed approaches  can be fitted into any deep network architecture, however for the evaluation we chose the Deeplab v2 network (without the post-processing based on CRFs) with ResNet-101 as feature extractor \cite{chen2018deeplab} pre-trained \cite{nekrasov} on the MSCOCO dataset \cite{lin2014microsoft}.
The pre-training of the feature extractor (as done also in other incremental learning works as \cite{li2018learning}) is needed since the Pascal VOC 2012 is too small to be used for training the Deeplab v2 from scratch. However MSCOCO data are used only for the initialization of the feature extractor and the contained labeling information, even if there are overlapping classes, is related to a different task (i.e., image classification). 

\begin{figure}
\centering
\includegraphics[trim={1.5cm 3.4cm 1.05cm 3.1cm}, clip, width=\linewidth]{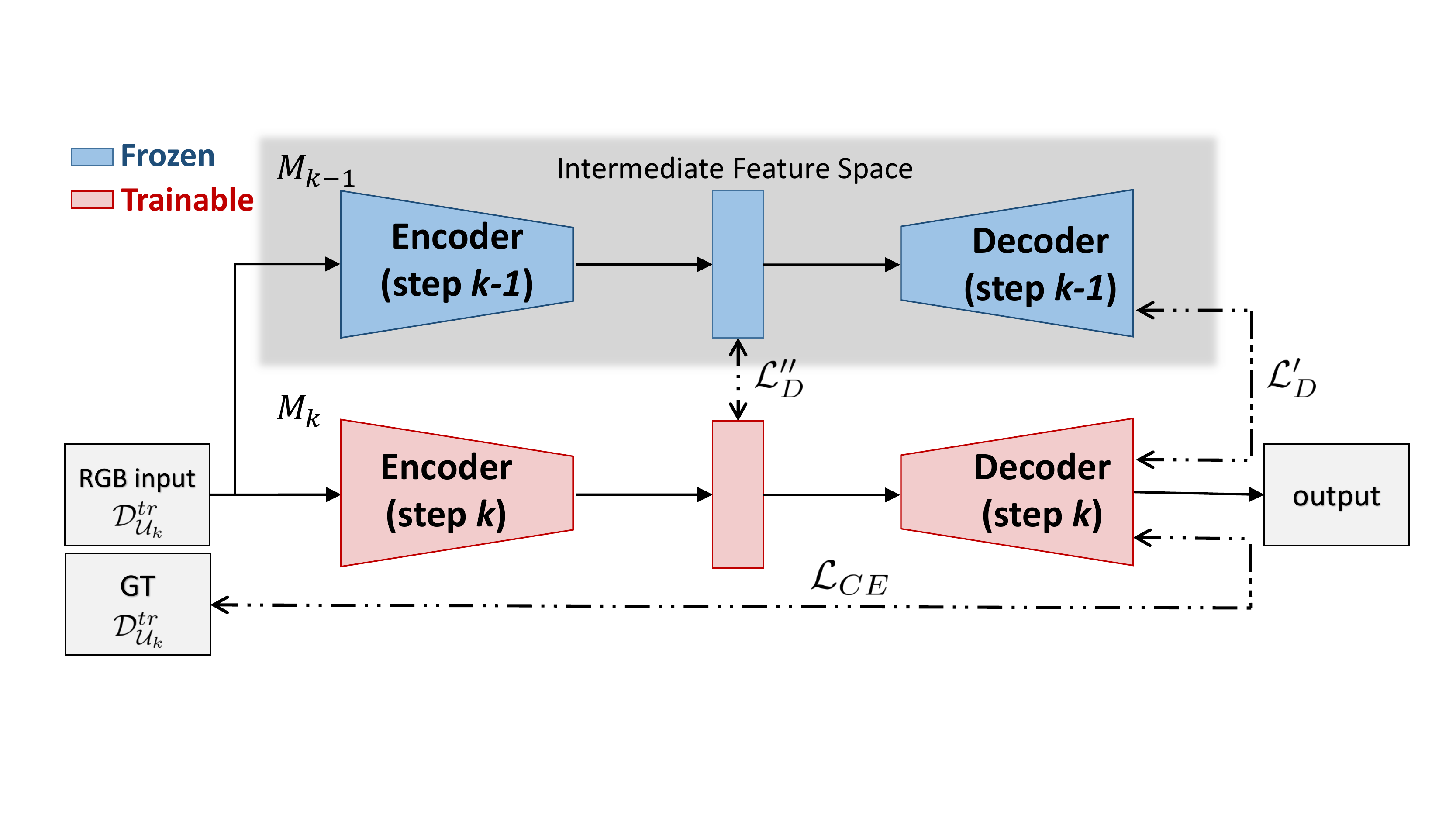}
\caption{Overview of the $k$-th incremental step of our learning framework for semantic segmentation of RGB images.}
\label{fig:architecture}
\vspace{-0.35cm}
\end{figure}

The various procedures to achieve incremental learning in semantic segmentation are now introduced: see  Fig.~\ref{fig:architecture} for a general overview of the approach.
We start by training the chosen network architecture in the first stage to recognize the classes in $\mathcal{S}_0$ with the corresponding training data $\mathcal{D}^{tr}_0$. 
The network is trained in a supervised way with a standard cross-entropy loss and after training we save the obtained model as $M_0$.
Then, we  perform a set of incremental steps indexed by $k=1,2,...$ to make the model learn every time a new set of classes $\mathcal{U}_k$. 
At the $k$-th incremental step, the current training set $\mathcal{D}_k^{tr}$ is built with images that contain samples from at least one of the new classes. Notice that they can possibly contain also pixels belonging to previously seen classes and of course the background class is present in almost all images. During step $k$, the model $M_{k-1}$ is loaded and updated exploiting  a linear combination of two losses: a cross-entropy loss $\mathcal{L}_{CE}$, which learns how  label the classes, and a distillation loss $\mathcal{L}_{D}$, which helps to retain knowledge of previously seen classes and will be detailed in the following. 
After the $k$-th incremental step, we save the current model as $M_k$ and the described procedure is repeated every time a new set of classes to be learned is taken into account.
The total loss $\mathcal{L}$ to train the model is:
\begin{equation}
\mathcal{L} = \mathcal{L}_{CE} + \lambda_D \mathcal{L}_D
\end{equation}

The parameter $\lambda_D$ balances the two terms. If we set $\lambda_D=0$ then we are considering the simplest scenario of fine-tuning in which no knowledge distillation is applied and the cross-entropy loss is applied to both unseen and seen classes (but in $\mathcal{D}_k^{tr}$ there is a large unbalance toward the new ones, see Section \ref{sec:problem}).  
As already pointed out, we expect this case to exhibit  catastrophic forgetting.

During the $k$-th incremental step the cross-entropy loss $\mathcal{L}_{CE}$ is applied to all the classes and it is defined as:

\begin{equation}
\label{eq:CE}
\mathcal{L}_{CE} = 
- \frac{1}{\lvert \mathcal{D}_{k}^{tr} \rvert}  \! 
\sum_{\mathbf{X}_n \in \mathcal{D}_{k}^{tr}} \!
\sum_{c \in \mathcal{S}_{k-1} \cup \mathcal{U}_k}
\mathbf{Y}_n [c] \! \cdot \!
\log \left( M_k \left(\mathbf{X}_n \right) [c]   \right)
\end{equation}
where $\mathbf{Y}_n [c]$ and $M_k \left(\mathbf{X}_n \right) [c]$ are respectively the one-hot encoded ground truth and the output of the  network  corresponding to the estimated score for class $c$. Notice that the sum is computed on both old and new classes because in practice old classes will continue to appear. However since the new classes are much more likely in $\mathcal{D}_k^{tr}$, there is a clear unbalance toward them leading to catastrophic forgetting \cite{wu2019}.
We introduce two possible strategies for defining the distillation loss $\mathcal{L}_{D}$ which only depend on the previous model $M_{k-1}$ avoiding the need for large storage. 
 
\subsection{Distillation on the Output Layer ($\mathcal{L}_{D}'$)}

The first considered distillation term $\mathcal{L}_D'$ for semantic segmentation is the masked cross-entropy loss between the logits produced by the output of the softmax layer in the previous model $M_{k-1}$ and the output of the softmax layer in the current model $M_{k}$ (assume that we  currently are at the $k$-th incremental step). The cross-entropy is masked to consider already seen classes only since we want to guide the learning process to retain them, i.e.:

\begin{equation}
\label{eq:D1}
\begin{aligned}
\mathcal{L}_{D}' \! = \!
- \frac{1}{\lvert \mathcal{D}_{k}^{tr} \rvert}  
 \! \sum_{\mathbf{X}_n \in \mathcal{D}_{k}^{tr}}
 \!  \sum_{c \in \mathcal{S}_{k-1}}
\! \! \! M_{k-1} \! \left(\mathbf{X}_n \!\right)\!  [c] \  \!\!\! \cdot  \! 
\log \left( M_k \! \left(\mathbf{X}_n \!\right) \! [c]   \right)
\end{aligned}
\end{equation}

\begin{figure}%
    \centering
    \subfloat[\scriptsize Encoder trainable]{{\includegraphics[trim={0cm 12cm 22.5cm 0cm}, clip, width=0.24\textwidth, valign=b]{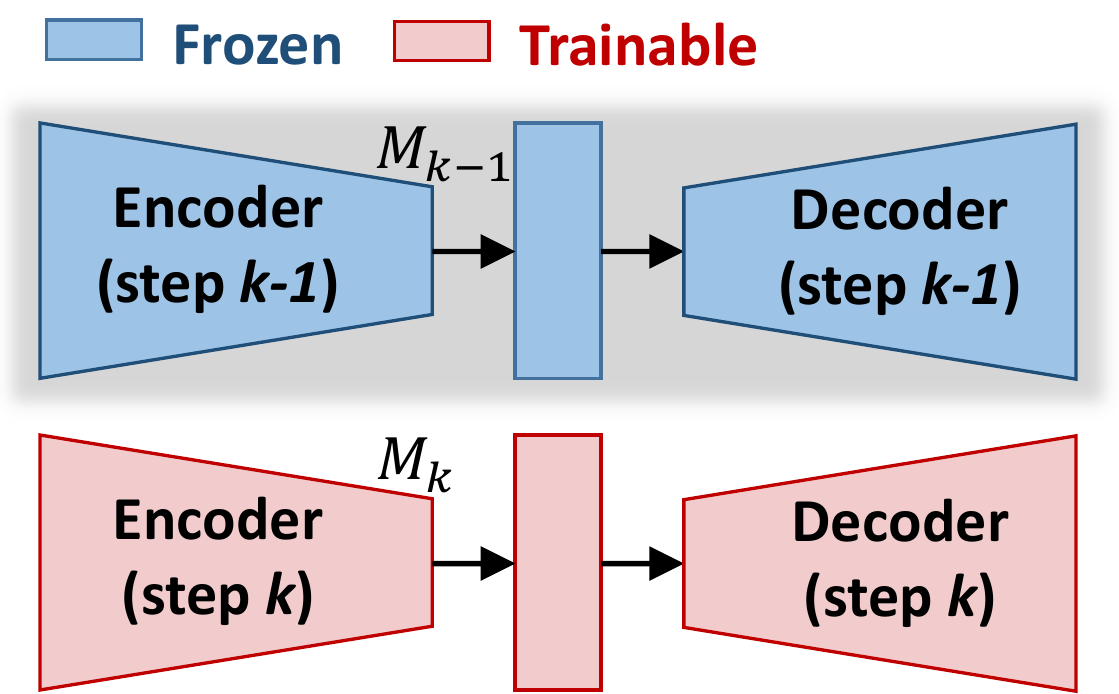} }}%
    \subfloat[\scriptsize Encoder frozen $E_F$]{{\includegraphics[trim={0cm 12cm 22.5cm 0cm}, clip, width=0.24\textwidth, valign=b]{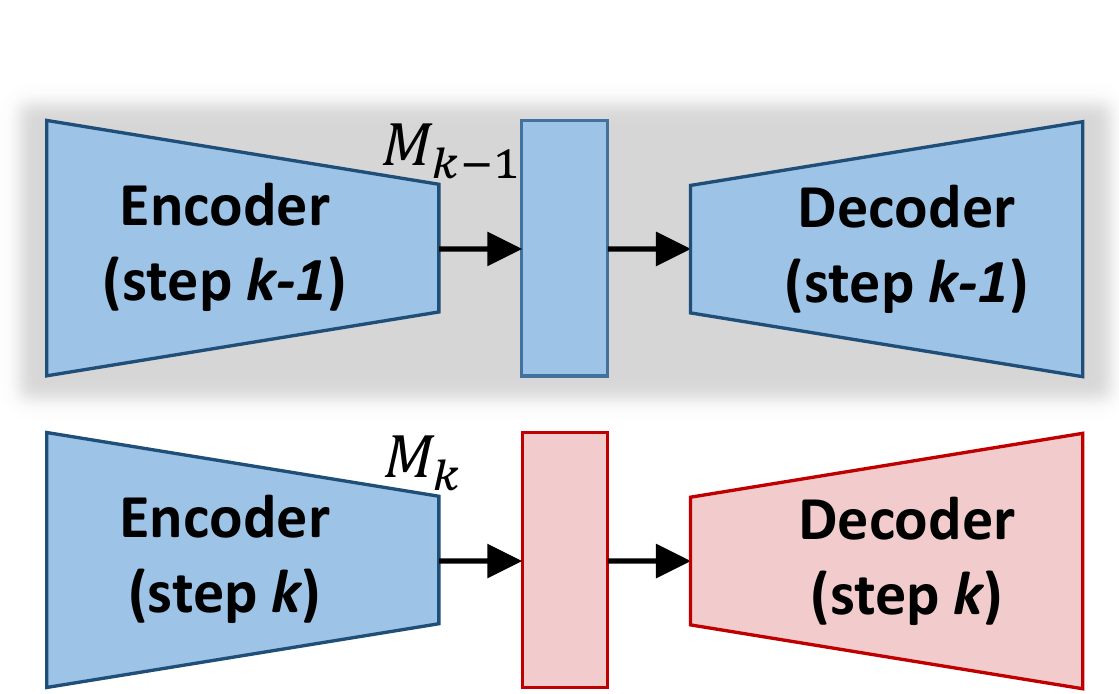} }}%
    \vspace{0.1cm}
    \caption{Freezing schemes of the encoder at $k$-th incremental step. The whole model at previous step, i.e. $M_{k-1}$, is always completely frozen and is employed only for knowledge distillation.}%
    \label{fig:encoders}%
\end{figure}

The loss $\mathcal{L}_{D}'$ is our baseline model and some enhancements of the scheme have been evaluated.
A first modification moves from the consideration that the encoder $E$ aims at extracting some intermediate feature representation from the input information: hence the encoder part of the network can be frozen to the status it reached after the previous steps ($E_F$ in short, see  Fig.~\ref{fig:encoders}). In this way the network is constrained to learn new classes only through the decoder, while preserving the features extraction capabilities unchanged from the previous training stage. We evaluated this approach both with and without the application of the distillation loss in Eq.~(\ref{eq:D1}). 

\subsection{Distillation on Intermediate Feature Space ($\mathcal{L}_{D}''$)}

A different approach we designed to preserve previous knowledge by keeping the encoder similar to the already learned model is to apply a knowledge distillation function to the intermediate level of the features space before the decoding stage. The distillation function on the features space in this case should be no longer the cross-entropy but rather the $L2$ loss. This choice is due to the fact that the considered layer is not anymore a classification layer but instead  just an internal stage where the output should be kept close to the previous one in, e.g., $L2$-norm.  
Empirically, we found that using cross-entropy or $L1$ lead to worse results.
Considering that model $M_k$ can be decomposed into an encoder $E_k$ and a decoder, the distillation term would become:

\begin{equation}
\label{eq:D2}
\mathcal{L}_{D}'' =  
\frac{\Vert  E_{k-1}(\mathbf{X}_n) - E_{k}(\mathbf{X}_n)   \Vert_2^2}
{\lvert \mathcal{D}_{k}^{tr} \rvert}
\end{equation}

where $E_k(\mathbf{X}_n)$ denotes the features computed by $E_k$ when a generic image $\mathbf{X}_n \in \mathcal{D}_{k}^{tr}$ is fed as input. 

A summary of the proposed strategies 
is shown in Fig.~\ref{fig:architecture} where the different losses are shown.
As a final remark, we also tried a combination of the described distillation losses but it did not provide relevant enhancements.

\section{Experimental Results}
\label{sec:results}

For the experimental evaluation we selected the Deeplab v2 architecture and we performed the tests on the Pascal VOC2012 \cite{pascalvoc2012} benchmark. This widely used dataset consists of $10582$ images in the training split and $1449$ in the validation split with a total of $21$ different classes (background included). Since the test set has not been made available, all the results have been computed on the validation split as done by most approaches in the literature.

We trained our network with Stochastic Gradient Descent (SGD) as done in \cite{chen2018deeplab}. The initial stage of training of the network on the set $\mathcal{S}_0$ is performed by setting the starting learning rate to $10^{-4}$ and training for $\lvert \mathcal{S}_0 \rvert \cdot 1000$ steps  decreasing the learning rate up to $10^{-6}$ with a polynomial decay rule with power $0.9$. 
We included weight decay regularization of $10^{-4}$ and we employed a batch size of $4$ images. 
The incremental training steps $k=1,2,...$ have been performed employing a lower learning rate to better preserve previous weights. In this case the learning rate starts from $5 \cdot 10^{-5}$ and decreases up to $10^{-6}$ after  $\lvert \mathcal{U}_k \rvert \cdot 1000$ steps of polynomial decay. Notice that we train the network for a number of steps which is proportional to the number of new classes to be learned. 
We used TensorFlow \cite{abadi2016tensorflow} to develop and train the network: the overall training procedure takes around 5 hours on a NVIDIA 2080 Ti GPU. The code is available online  at \url{https://lttm.dei.unipd.it/paper_data/IL}.
The metrics we considered are the most widely used for semantic segmentation: 
the per-class Intersection over Union (IoU), the mean Pixel Accuracy (mPA), the mean Class Accuracy (mCA) and the mean IoU (mIoU) \cite{csurka2013good}.

\begin{table*}[htbp]
{\footnotesize
\setlength{\tabcolsep}{1.6pt}
\centering
\begin{tabular}{|c|cccccccccccccccccccc:c|c|ccc|}
\hline
$M_1 (20)$ & \rotatebox{90}{backgr.} &  \rotatebox{90}{aero} &  \rotatebox{90}{bike} &  \rotatebox{90}{bird} &\rotatebox{90}{boat} & \rotatebox{90}{bottle} & \rotatebox{90}{bus} 
  &\rotatebox{90}{car} & \rotatebox{90}{cat} & \rotatebox{90}{chair} & \rotatebox{90}{cow} & \rotatebox{90}{din. table}& \rotatebox{90}{dog} & \rotatebox{90}{horse} 
  & \rotatebox{90}{mbike} & \rotatebox{90}{person} & \rotatebox{90}{plant} &  \rotatebox{90}{sheep} & \rotatebox{90}{sofa} & \rotatebox{90}{train} & \rotatebox{90}{\textbf{mIoU old}} & \rotatebox{90}{tv} & \rotatebox{90}{\textbf{mIoU}} & \rotatebox{90}{\textbf{mPA}} & \rotatebox{90}{\textbf{mCA}}\\
 \hline

Fine-tuning & 90.2 & 80.8 & 33.3 & 83.1 & 53.7 & 68.2 & 84.6 & 78.0 & 83.2 & 32.1 & 73.4 & 52.6 & 76.6 & 72.7 & 68.8 & 79.8 & 43.8 & 76.5 & 46.5 & 68.4 & 67.3 & 20.1 & 65.1 & 90.7 & 76.5 \\

$\mathcal{L}_{D}'$ & 92.0 & 83.9 & 37.0 & 84.0 & 58.8 & 70.9 & 90.9 & 82.5 & 86.1 & 32.1 & 72.5 & 51.0 & 79.9 & 72.3 & 77.3 & 80.9 & 45.1 & 78.1 & 45.7 & 79.9 & 70.0 & 35.3 & 68.4 & 92.5 & 79.5 \\

$E_F$ & 92.7 & 86.2 & 32.6 & 82.9 & 61.7 & 74.6 & 92.9 & 83.1 & 87.7 & 27.4 & 79.4 & 59.0 & 79.4 & 76.9 & 77.2 & 81.2 & 49.6 & 80.8 & 49.3 & 83.4 & 71.9 & 43.3 & 70.5 & 93.2 & 81.4 \\

$E_F$, $\mathcal{L}_{D}'$ & 92.9 & 86.1 & 37.1 & 83.6 & 62.2 & 76.1 & 93.2 & 82.9 & 88.3 & 30.6 & 79.6 & 58.5 & 80.3 & 77.6 & 77.2 & 81.8 & 49.8 & 81.0 & 47.0 & 84.5 & 72.5 & \textbf{51.4} & 71.5 & 93.4 & 82.5 \\


$\mathcal{L}_{D}''$ & 92.9 & 84.8 & 36.4 & 82.6 & 63.5 & 75.0 & 92.2 & 83.6 & 88.3 & 29.5 & 80.3 & 59.6 & 79.7 & 80.2 & 78.9 & 81.2 & 49.7 & 78.9 & 51.0 & 84.1 & 72.6 & 50.6 & 71.6 & 93.4 & \textbf{83.4} \\

$\mathcal{L}_{D}', \mathcal{L}_{D}''$ & 92.9 & 86.0 & 36.5 & 84.4 & 61.8 & 76.2 & 93.1 & 83.1 & 88.6 & 30.4 & 79.7 & 58.7 & 80.4 & 78.1 & 76.4 & 82.0 & 50.5 & 81.0 & 50.4 & 85.1 & \textbf{72.8} & 49.9 & \textbf{71.7} & \textbf{93.5} & \textbf{83.4} \\


 \hline

$M_0(0-19)$ & 93.4 & 85.5 & 37.1 & 86.2 & 62.2 & 77.9 & 93.4 & 83.5 & 89.3 & 32.6 & 80.7 & 57.3 & 81.5 & 81.2 & 77.7 & 83.0 & 51.5 & 81.6 & 48.2 & 85.0 & 73.4 & -  & 73.4 & 93.9 & 84.3 \\

$M_0(0-20)$ & 93.4 & 85.4 & 36.7 & 85.7 & 63.3 & 78.7 & 92.7 & 82.4 & 89.7 & 35.4 & 80.9 & 52.9 & 82.4 & 82.0 & 76.8 & 83.6 & 52.3 & 82.4 & 51.1 & 86.4 & 73.7 & 70.5 & 73.6 & 93.9 & 84.2 \\
\hline
\end{tabular}
}
\caption{Per-class IoU on the Pascal VOC2012 under some settings when the last class, i.e. the tv/monitor class, is added.}
\label{tab:pascal_0_19_20}
\end{table*}
\vspace{0.4cm}

\subsection{Addition of One Class}
\label{subsec:single}

Following \cite{shmelkov2017incremental} we first analyze the addition of the last class, in alphabetical order, to our network. Specifically, we consider $\mathcal{S}_0 = \lbrace  c_0,c_1,...,c_{19}  \rbrace$ and $\mathcal{U}_1 = \lbrace  c_{20}  \rbrace = \lbrace \mathit{tv \backslash monitor}  \rbrace$. 
A summary of the evaluation of the proposed methodologies  on the VOC2012 validation split is reported in Table~\ref{tab:pascal_0_19_20}. We indicate as $M_0 (0-19)$ the first standard training of the network using $\mathcal{D}^{tr}_0$ as training dataset. The network is then updated exploiting the dataset $\mathcal{D}^{tr}_1$ and the resulting model is referred to as $M_1 (20)$.
From the first row of Table~\ref{tab:pascal_0_19_20} we can appreciate that fine-tuning the network leads to an evident degradation of the performance with a final mIoU of $65.1\%$. This is a clear confirmation of the catastrophic forgetting phenomenon in the semantic segmentation scenario even with the addition of just one single class. 
The reference model, indeed, where all the $21$ classes are learned at once (we call it $M_0 (0-20)$) achieves a mIoU of $73.6\%$. The main issue of the fine-tuning approach is that it predicts too frequently the last  class, 
 as proved by the fact that the model has a very high pixel accuracy for the $\mathit{tv/monitor}$ class 
but a very poor IoU of $20.1\%$. 
This is due to the high number of false positive detection of the considered class which are not taken into account by the pixel accuracy measure. 
On the same class, the proposed methods are all able to outperform the fine-tuning approach in terms of IoU by large margin.
Knowledge distillation strategies and the procedure of freezing the encoder provide better results because they act as regularization constraints. Interestingly those procedures allow to achieve higher accuracy not only on previously learned classes but also on newly added ones, which might be unexpected if we do not consider the regularization behavior of those terms. 
We can appreciate that the distillation on the output $\mathcal{L}_{D}'$ alone is able to improve the average mIoU by $3.3\%$ with respect to the standard case. Furthermore it leads to a much better IoU on the new class, greatly reducing the aforementioned false positives issue. If we completely freeze the encoder $E$ without applying knowledge distillation the model improves the mIoU by $5.4\%$. 
If we combine the two mentioned approaches, i.e. we freeze $E$ and we apply $\mathcal{L}_{D}'$ as distillation loss, the mIoU further improves to  $71.5\%$ with an  improvement of $6.4\%$, higher than each of the two methods alone (also the performance on the new class is higher).

If we apply a $L2$ loss at the intermediate features space, i.e., to use $\mathcal{L}_{D}''$, the model achieves $71.6\%$ of mIoU, which is $6.5\%$ higher than the standard approach.
It is noticeable that two completely different approaches  to preserve knowledge from the previous model, namely ``$M_1(20)$ with $E_F$, $\mathcal{L}_{D}'$" (which applies a cross-entropy between the outputs with encoder frozen) and ``$M_1(20)$ with $\mathcal{L}_{D}''$" (which applies a $L2$-loss between features spaces), achieve similar and high results both on the new class and on old ones. Notice that if the encoder is frozen then it does not make sense to enable the $\mathcal{L}_{D}''$ loss.

An interesting aspect is that the changes in performance on previously seen classes are correlated with the class being added. Some classes have even higher results in terms of mIoU than before because their prediction has been reinforced through the new training set. For example, objects of the classes $\mathit{sofa}$ or $\mathit{dining\ table}$ are typically present in scenes containing a $\mathit{tv/monitor}$. 
Classes containing uncorrelated objects that are not present inside the new set of samples $\mathcal{D}^{tr}_1$ instead get more easily lost, for example the $\mathit{bird}$ or $\mathit{horse}$ which are not present in indoor scenes typically associated  with the $tv/monitor$ class being added.

\subsection{Addition of Five Classes}

In this section we tackle a more challenging scenario where the initial learning stage is followed by one step of incremental learning with the last $5$ classes to learn. 
\begin{table*}[h]
{
\footnotesize
\setlength{\tabcolsep}{1.6pt}
\centering
\begin{tabular}{|c|cccccccccccccccc:c|ccccc:c|ccc|}
\hline
 $M_1 (16-20)$ & \rotatebox{90}{backgr.} &  \rotatebox{90}{aero} &  \rotatebox{90}{bike} &  \rotatebox{90}{bird} &\rotatebox{90}{boat} & \rotatebox{90}{bottle} & \rotatebox{90}{bus} 
  &\rotatebox{90}{car} & \rotatebox{90}{cat} & \rotatebox{90}{chair} & \rotatebox{90}{cow} & \rotatebox{90}{din. table}& \rotatebox{90}{dog} & \rotatebox{90}{horse} 
  & \rotatebox{90}{mbike} & \rotatebox{90}{person} & \rotatebox{90}{\textbf{mIoU old}} & \rotatebox{90}{plant} &  \rotatebox{90}{sheep} & \rotatebox{90}{sofa} & \rotatebox{90}{train} & \rotatebox{90}{tv} & \rotatebox{90}{\textbf{mIoU new}} & \rotatebox{90}{\textbf{mIoU}} & \rotatebox{90}{\textbf{mPA}} & \rotatebox{90}{\textbf{mCA}}\\
 \hline

Fine-tuning & 89.7 & 59.5 & 34.6 & 68.2 & 58.1 & 58.8 & 59.2 & 79.2 & 80.2 & 30.0 & 12.7 & 51.0 & 72.5 & 61.7 & 74.4 & 79.4 & 60.6 & 36.4 & 32.4 & 27.2 & 55.2 & 42.4 & 38.7 & 55.4 & 88.4 & 70.6 \\

$\mathcal{L}_{D}'$ & 91.4 & 85.0 & 35.6 & 84.8 & 61.8 & 70.5 & 85.6 & 77.9 & 83.6 & 30.7 & 72.0 & 45.4 & 76.1 & 76.9 & 77.0 & 81.3 & \textbf{71.0} & 33.8 & 54.9 & 30.8 & 73.9 & 51.6 & \textbf{49.0} & \textbf{65.7} & \textbf{91.6} & \textbf{78.0}\\

$E_F$, $\mathcal{L}_{D}'$ & 91.7 & 83.4 & 35.6 & 78.7 & 60.9 & 73.0 & 65.8 & 82.2 & 87.0 & 30.2 & 58.0 & 55.3 & 80.0 & 78.3 & 78.5 & 81.4 & 70.0 & 35.3 & 46.1 & 32.3 & 62.1 & 53.5 & 45.8 & 64.2 & 91.5 & 76.1 \\


$\mathcal{L}_{D}''$ & 90.9 & 81.4 & 33.9 & 80.3 & 61.9 & 67.4 & 73.1 & 81.8 & 84.8 & 31.3 & 0.4 & 55.8 & 76.1 & 72.2 & 77.7 & 81.2 & 65.6 & 39.4 & 31.8 & 31.3 & 64.1 & 52.9  & 43.9 & 60.5 & 90.0 & 74.9\\ \hline


$M_0(0-15)$ & 94.0 & 83.5 & 36.1 & 85.5 & 61.0 & 77.7 & 94.1 & 82.8 & 90.0 & 40.0 & 82.8 & 54.9 & 83.4 & 81.2 & 78.3 & 83.2 & 75.5 & - & - & - & - & - & - & 75.5 & 94.6 & 86.4\\

$M_0(0-20)$ & 93.4 & 85.4 & 36.7 & 85.7 & 63.3 & 78.7 & 92.7 & 82.4 & 89.7 & 35.4 & 80.9 & 52.9 & 82.4 & 82.0 & 76.8 & 83.6 & 75.1 & 52.3 & 82.4 & 51.1 & 86.4 & 70.5 & 68.5 & 73.6 & 93.9 & 84.2 \\
\hline
\end{tabular}
}
\vspace{-0.2cm}
\caption{Per-class IoU on the Pascal VOC2012 under some settings when $5$ classes are added all at once.}
\label{tab:pascal_0_15_20}
\vspace{-0.08cm}
\end{table*}
First, the addition of the last $5$ classes at once (referred to as $15-20$) is discussed and the results are shown in Table~\ref{tab:pascal_0_15_20}. In this setting the results are much lower than in the previous cases where a single class was added at a time since there is a larger amount of information to be learned.
In particular, the fine-tuning approach exhibits an even larger drop in accuracy because it overestimates the presence of the new classes. 
We can confirm this by looking at the IoU scores of the newly added classes which are often lower than the proposed approaches by a large margin. 
In this setting the distillation on the output layer, ``$M_1(16-20)$ with $\mathcal{L}_{D}'$", achieves the highest accuracy. In general in this case the approaches based on $\mathcal{L}_{D}'$ outperform the other ones. 
It is interesting to notice that some previously seen classes exhibit a clear catastrophic forgetting phenomenon because the updated models mislead them with visually similar classes belonging to the set of new classes. This is particularly true, for example, for the $\mathit{cow}$ and $\mathit{chair}$ classes which are often misled (low IoU and low pixel accuracy for these classes) with the newly added classes $\mathit{sheep}$ and $\mathit{sofa}$ that have similar shapes (low IoU but high pixel accuracy for these classes). This can be seen also in the qualitative results  in Fig.~\ref{fig:addition_of_5_at_once}.
For example, in the first two rows the $\mathit{tv/monitor}$ and the $\mathit{sofa}$ classes (which are added during the incremental step) are erroneously predicted in  the $\mathit{background}$ region, while these classes are correctly handled by applying  $\mathcal{L}_D'$ and freezing the encoder. Additionally, in the third row the na\"\i ve approach predicts the $\mathit{person}$ class in spite of the $\mathit{sofa}$ while this artifact is not present when using $\mathcal{L}_{D}'$.


\begin{figure}[tb]{}

\setlength\tabcolsep{1pt} 
\subfloat{
\begin{tabular}{ccccc} 
  \scriptsize RGB & 
  \scriptsize GT & 
  \scriptsize Fine-tuning & 
  \scriptsize $M_1 E_F \mathcal{L}_{D}'$ &  
  \scriptsize $M_0$ \\ 
  
  \includegraphics[width=0.185\linewidth]{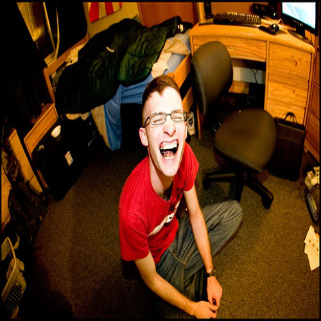} &
  \includegraphics[width=0.185\linewidth]{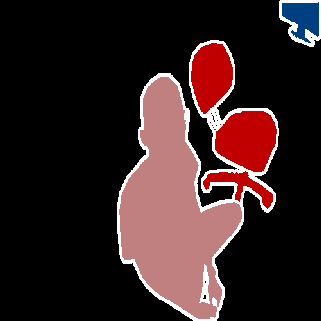} &
  \includegraphics[width=0.185\linewidth]{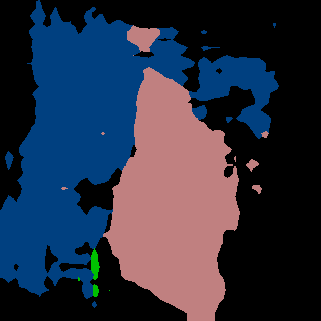} &
  \includegraphics[width=0.185\linewidth]{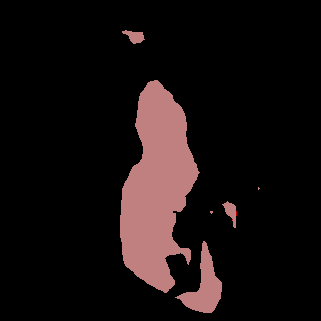} &
  \includegraphics[width=0.185\linewidth]{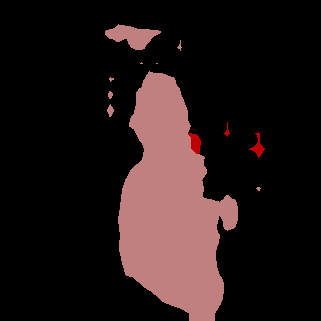} \\
  
  \includegraphics[width=0.185\linewidth]{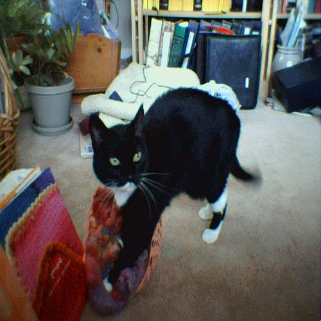}&
  \includegraphics[width=0.185\linewidth]{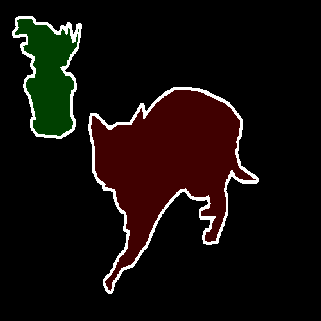} &
  \includegraphics[width=0.185\linewidth]{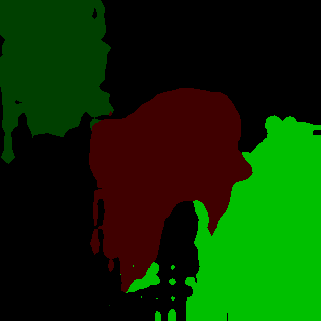}  &
  \includegraphics[width=0.185\linewidth]{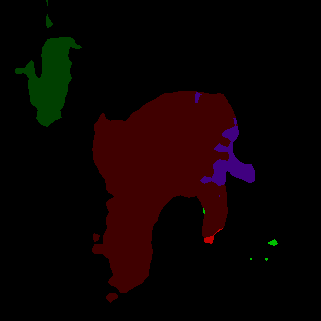} &
  \includegraphics[width=0.185\linewidth]{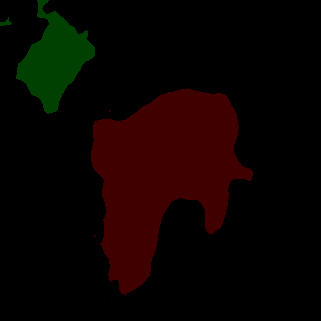} \\
  
   \includegraphics[width=0.185\linewidth]{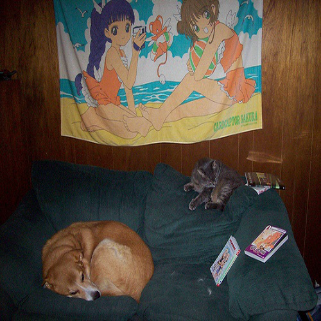} & 
   \includegraphics[width=0.185\linewidth]{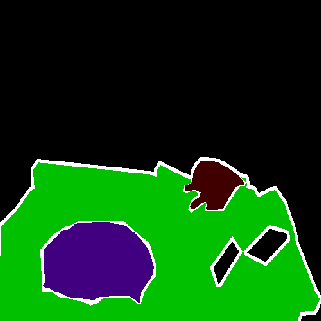} &
   \includegraphics[width=0.185\linewidth]{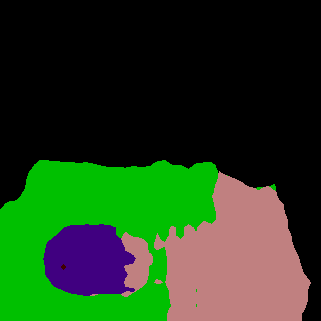} &
   \includegraphics[width=0.185\linewidth]{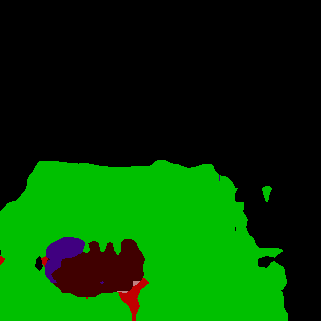}  &
   \includegraphics[width=0.185\linewidth]{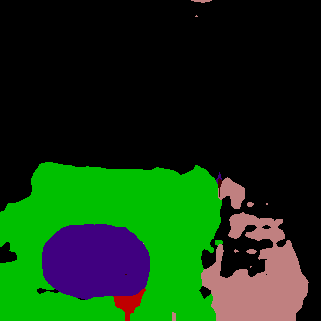} \\
 \end{tabular}
}
\centering
\vspace{-0.75cm}
\subfloat{
\hspace{0.3cm}
\resizebox{6.5cm}{!}{%
\begin{tabular}{ccccccccc}
& & & & & & & & \\ \cline{9-9}
\cellcolor[HTML]{000000}{\color[HTML]{FFFFFF} \textbf{background}} & 
\cellcolor[HTML]{400000}{\color[HTML]{FFFFFF} \textbf{cat}} & 
\cellcolor[HTML]{C00000}{\color[HTML]{FFFFFF} \textbf{chair}} & 
\cellcolor[HTML]{400080}{\color[HTML]{FFFFFF} \textbf{dog}} & 
\cellcolor[HTML]{C08080}\textbf{person} & 
\cellcolor[HTML]{004000}{\color[HTML]{FFFFFF} \textbf{plant}} & 
\cellcolor[HTML]{00C000}\textbf{sofa} & 
\multicolumn{1}{c|}{\cellcolor[HTML]{004080}{\color[HTML]{FFFFFF} \textbf{tv}}} &
\multicolumn{1}{c|}{\textbf{unlabeled}}\\ \cline{9-9} 

\end{tabular}%
}
}%
\caption{Qualitative results on sample scenes for the addition of five classes all at once (\textit{best viewed in colors}).}
\label{fig:addition_of_5_at_once}
\end{figure}


\begin{table*}[htbp]
{
\footnotesize
\setlength{\tabcolsep}{1.43pt}
\centering
\begin{tabular}{|c|cccccccccccccccc:c|c:c:c:c:c:c|ccc|}
\hline
 $M_5 (16\rightarrow20)$ & \rotatebox{90}{backgr.} &  \rotatebox{90}{aero} &  \rotatebox{90}{bike} &  \rotatebox{90}{bird} &\rotatebox{90}{boat} & \rotatebox{90}{bottle} & \rotatebox{90}{bus} 
  &\rotatebox{90}{car} & \rotatebox{90}{cat} & \rotatebox{90}{chair} & \rotatebox{90}{cow} & \rotatebox{90}{din. table}& \rotatebox{90}{dog} & \rotatebox{90}{horse} 
  & \rotatebox{90}{mbike} & \rotatebox{90}{person} & \rotatebox{90}{\textbf{mIoU old}} & \rotatebox{90}{plant} &  \rotatebox{90}{sheep} & \rotatebox{90}{sofa} & \rotatebox{90}{train} & \rotatebox{90}{tv} & \rotatebox{90}{\textbf{mIoU new}} & \rotatebox{90}{\textbf{mIoU}} & \rotatebox{90}{\textbf{mPA}} & \rotatebox{90}{\textbf{mCA}}\\
 \hline

Fine-tuning & 87.9 & 25.6 & 29.0 & 51.2 & 1.7 & 57.8 & 10.5 & 64.8 & 80.5 & 30.8 & 22.9 & 52.7 & 66.8 & 52.1 & 51.9 & 78.1 & 47.8 & \textbf{36.5} & 44.7 & \textbf{31.8} & 35.1 & 17.1 & 33.0 & 44.2 & 86.1 & 55.7 \\

$\mathcal{L}_{D}'$ & 89.7 & 51.2 & 29.7 & 77.5 & 15.0 & 62.7 & 29.1 & 78.5 & 75.7 & 24.4 & 55.6 & 44.8 & 76.2 & 62.5 & 65.6 & 80.1 & 57.4 & 25.5 & 35.7 & 30.8 & 42.3 & 40.4 & 34.9 & 52.0 & 88.6 & 63.2 \\

$E_F$, $\mathcal{L}_{D}'$ & 91.1 & 73.9 & 31.9 & 81.4 & 59.5 & 71.9 & 73.1 & 82.1 & 87.1 & 27.2 & 77.4 & 56.4 & 79.1 & 79.9 & 76.1 & 80.7 & \textbf{70.5} & 31.6 & 55.3 & 30.4 & \textbf{62.2} & \textbf{41.4} & 44.2 & \textbf{64.3} & \textbf{91.3} & \textbf{75.2} \\



$\mathcal{L}_{D}''$ & 90.3 & 54.2 & 28.2 & 78.4 & 52.5 & 69.8 & 59.5 & 78.5 & 86.3 & 28.8 & 72.3 & 57.4 & 76.3 & 77.1 & 65.8 & 79.3 & 65.9 & 36.3 & \textbf{65.5} & 31.6 & 54.7 & 38.9 & \textbf{45.4} & 61.0 & 90.4 & 71.0 \\\hline


$M_0(0-15)$ & 94.0 & 83.5 & 36.1 & 85.5 & 61.0 & 77.7 & 94.1 & 82.8 & 90.0 & 40.0 & 82.8 & 54.9 & 83.4 & 81.2 & 78.3 & 83.2 & 75.5 & - & - & - & - & - & - & 75.5 & 94.6 & 86.4 \\

$M_0(0-20)$ & 93.4 & 85.4 & 36.7 & 85.7 & 63.3 & 78.7 & 92.7 & 82.4 & 89.7 & 35.4 & 80.9 & 52.9 & 82.4 & 82.0 & 76.8 & 83.6 & 75.1 & 52.3 & 82.4 & 51.1 & 86.4 & 70.5 & 68.5 & 73.6 & 93.9 & 84.2 \\
\hline
\end{tabular}
\caption{Per-class IoU on the Pascal VOC2012 under some settings when $5$ classes are added sequentially.}
\label{tab:pascal_0_15_16_17_18_19_20}
}
\end{table*}

The last experiment presented here is the one in which the last $5$ classes are progressively added one by one: the final model is referred to as $M_5 (16 \rightarrow 20)$. The results are reported in Table~\ref{tab:pascal_0_15_16_17_18_19_20} where we can appreciate a large gain of $20\%$ of mIoU between the best proposed method (i.e., ``$M_5 (16 \rightarrow 20)$ with $E_F, \mathcal{L}_{D}'$") and the standard approach. In this case freezing the encoder and distilling the knowledge is the best approach because the addition of one single class do not alter too much the responses of the whole network: distilling the knowledge from the previous model when the encoder is fixed guides the decoder to modify only the responses for the new class. 

\begin{table}[]
{
\footnotesize
 \vspace{0.1cm} 
\setlength{\tabcolsep}{0.58pt}
\centering
\begin{tabular}{l|c|c|c||c|c|c||c|c|c||c|c|c|c|c|c|}
\multicolumn{1}{c}{} & \multicolumn{3}{c}{Fine-tuning} & \multicolumn{3}{c}{$\mathcal{L}_D'$} & \multicolumn{3}{c}{$E_F$, $\mathcal{L}_D'$} &
\multicolumn{3}{c}{$\mathcal{L}_D''$}  \vspace{0.045cm} \\ 
  \hhline{~|---||---||---||---|}  \vspace{-0.045cm} 
  & \scriptsize mIoU    & \scriptsize mPA   & \scriptsize mCA   & \scriptsize mIoU    & \scriptsize mPA   & \scriptsize mCA   & \scriptsize mIoU   & \scriptsize mPA   & \scriptsize  mCA  & 
  \scriptsize mIoU   & \scriptsize mPA   & \scriptsize mCA  \\
   \hhline{~|---||---||---||---|}
$M_1(16)$ & 71.2 & 93.7 & 82.5 & 72.4 & 94.2 & 83.0 & 72.5 & 94.1 & 83.5 & 
 72.2 & 93.9 & 84.3 \\
$M_2(17)$  &  53.8 & 90.0 & 61.8 & 68.1 & 93.4 & 78.5 & 68.4 & 93.3 & 79.5 & 
  60.0 & 91.6 & 69.4 \\
$M_3(18)$ & 57.7 & 87.7 & 68.7 & 63.3 & 90.8 & 74.5 & 66.5 & 91.5 & 79.4 & 
  65.5 & 90.7 & 76.8 \\
$M_4(19)$ &  39.3 & 85.9 & 47.4 & 54.1 & 89.2 & 64.3 & 61.3 & 90.6 & 72.5 & 
  52.1 & 89.0 & 60.6 \\
$M_5(20)$&  44.2 & 86.1 & 55.7 & 52.0 & 88.6 & 63.2 & 64.3 & 91.3 & 75.2 & 
61.0 & 90.4 & 71.0 \\
 \hhline{~|---||---||---||---|}
\end{tabular}
}
\caption{Mean IoU, mPA and mCA on the Pascal VOC2012 under some settings when $5$ classes are added sequentially.}
\label{tab:pascal_0_15_16_17_18_19_20_methods_vs_steps}
 \vspace{-0.2cm} 
\end{table}

The evolution of the models' mean performance over time is reported in Table~\ref{tab:pascal_0_15_16_17_18_19_20_methods_vs_steps} where the distribution of the drop of performance during the different steps is analyzed. In particular we can notice how the accuracy drop is affected by the specific class being added. As expected the larger drop is experienced when the classes $\mathit{sheep}$ or $\mathit{train}$ are added (models $M_2(17)$ and $M_4(19)$) because such classes 
are only sparsely correlated with  other classes (they mainly appear alone or with the $\mathit{person}$ class). 


\section{Conclusion and Future Work}
\label{sec:conclusion}

In this work we formally introduced the problem of incremental learning for semantic segmentation. A couple of novel distillation loss functions have been designed ad-hoc for the task. They have been combined with a cross-entropy loss and with the idea of freezing the encoder module to optimize the performance on new classes while preserving old ones.
Our method does not need any stored image of previous datasets 
and only the previous model is used to update the current one thus reducing memory consumption.

Experiments on the Pascal VOC2012 dataset show that the proposed methods were able to largely outperform the standard fine-tuning approach, thus alleviating the catastrophic forgetting phenomenon. However, the problem of incremental learning for semantic segmentation is a novel challenging task that needs advanced strategies to be tackled. This is proved by the fact that the results are lower than the ones achieved by the same architecture after a one-step training, i.e., when all training examples are available and employed at the same time.
In the future we plan to expand our set of experiments, to develop novel incremental learning strategies and to employ GANs to generate images containing already seen classes. Finally we will consider the scenario in which classes that will appear in the future are present from the beginning but labeled as background.

{\small
\bibliographystyle{ieee}
\bibliography{strings,refs}

\begin{thebibliography}{10}\itemsep=-1pt

\bibitem{abadi2016tensorflow}
M.~Abadi, P.~Barham, J.~Chen, Z.~Chen, A.~Davis, J.~Dean, M.~Devin,
  S.~Ghemawat, G.~Irving, M.~Isard, et~al.
\newblock Tensorflow: A system for large-scale machine learning.
\newblock In {\em 12th Symposium on Operating Systems Design and
  Implementation}, pages 265--283, 2016.

\bibitem{aljundi2018memory}
R.~Aljundi, F.~Babiloni, M.~Elhoseiny, M.~Rohrbach, and T.~Tuytelaars.
\newblock Memory aware synapses: Learning what (not) to forget.
\newblock In {\em Proceedings of European Conference on Computer Vision
  (ECCV)}, pages 139--154, 2018.

\bibitem{biasetton2019unsupervised}
M.~Biasetton, U.~Michieli, G.~Agresti, and P.~Zanuttigh.
\newblock {Unsupervised Domain Adaptation for Semantic Segmentation of Urban
  Scenes}.
\newblock {\em Proceedings of IEEE Conference on Computer Vision and Pattern
  Recognition Workshops (CVPRW)}, 2019.

\bibitem{bucilua2006model}
C.~Buciluǎ, R.~Caruana, and A.~Niculescu-Mizil.
\newblock Model compression.
\newblock In {\em Proceedings of the 12th ACM International Conference on
  Knowledge Discovery and Data Mining (SIGKDD)}, pages 535--541. ACM, 2006.

\bibitem{castro2018end}
F.~M. Castro, M.~J. Mar{\'\i}n-Jim{\'e}nez, N.~Guil, C.~Schmid, and K.~Alahari.
\newblock End-to-end incremental learning.
\newblock In {\em Proceedings of European Conference on Computer Vision
  (ECCV)}, pages 233--248, 2018.

\bibitem{cauwenberghs2001incremental}
G.~Cauwenberghs and T.~Poggio.
\newblock Incremental and decremental support vector machine learning.
\newblock In {\em Advances in Neural Information Processing Systems (NIPS)},
  pages 409--415, 2001.

\bibitem{chaudhry2018riemannian}
A.~Chaudhry, P.~K. Dokania, T.~Ajanthan, and P.~H. Torr.
\newblock Riemannian walk for incremental learning: Understanding forgetting
  and intransigence.
\newblock In {\em Proceedings of European Conference on Computer Vision
  (ECCV)}, pages 532--547, 2018.

\bibitem{chen2018deeplab}
L.-C. Chen, G.~Papandreou, I.~Kokkinos, K.~Murphy, and A.~L. Yuille.
\newblock Deeplab: Semantic image segmentation with deep convolutional nets,
  atrous convolution, and fully connected crfs.
\newblock {\em IEEE Transactions on Pattern Analysis and Machine Intelligence
  (PAMI)}, 40(4):834--848, 2018.

\bibitem{csurka2013good}
G.~Csurka, D.~Larlus, F.~Perronnin, and F.~Meylan.
\newblock What is a good evaluation measure for semantic segmentation?
\newblock In {\em BMVC}, volume~27, page 2013. Citeseer, 2013.

\bibitem{pascalvoc2012}
M.~Everingham, L.~Van~Gool, C.~K.~I. Williams, J.~Winn, and A.~Zisserman.
\newblock {The {PASCAL} {V}isual {O}bject {C}lasses {C}hallenge 2012
  {(VOC2012)} {R}esults"}.
\newblock
  http://www.pascal-network.org/challenges/VOC/voc2012/workshop/index.html.

\bibitem{french1999catastrophic}
R.~M. French.
\newblock Catastrophic forgetting in connectionist networks.
\newblock {\em Trends in Cognitive Sciences}, 3(4):128--135, 1999.

\bibitem{furlanello2016active}
T.~Furlanello, J.~Zhao, A.~M. Saxe, L.~Itti, and B.~S. Tjan.
\newblock Active long term memory networks.
\newblock {\em arXiv preprint arXiv:1606.02355}, 2016.

\bibitem{goodfellow2013empirical}
I.~J. Goodfellow, M.~Mirza, D.~Xiao, A.~Courville, and Y.~Bengio.
\newblock An empirical investigation of catastrophic forgetting in
  gradient-based neural networks.
\newblock {\em International Conference on Learning Representations (ICLR)},
  2014.

\bibitem{hinton2015distilling}
G.~Hinton, O.~Vinyals, and J.~Dean.
\newblock Distilling the knowledge in a neural network.
\newblock {\em Neural Information Processing Systems (NIPS) Deep Learning and
  Representation Learning Workshop}, 2015.

\bibitem{hou2018lifelong}
S.~Hou, X.~Pan, C.~Change~Loy, Z.~Wang, and D.~Lin.
\newblock Lifelong learning via progressive distillation and retrospection.
\newblock In {\em Proceedings of European Conference on Computer Vision
  (ECCV)}, pages 437--452, 2018.

\bibitem{istrate2018incremental}
R.~Istrate, A.~C.~I. Malossi, C.~Bekas, and D.~Nikolopoulos.
\newblock Incremental training of deep convolutional neural networks.
\newblock {\em arXiv preprint arXiv:1803.10232}, 2018.

\bibitem{kirkpatrick2017overcoming}
J.~Kirkpatrick, R.~Pascanu, N.~Rabinowitz, J.~Veness, G.~Desjardins, A.~A.
  Rusu, K.~Milan, J.~Quan, T.~Ramalho, A.~Grabska-Barwinska, D.~Hassabis,
  C.~Clopath, D.~Kumaran, and R.~Hadsell.
\newblock Overcoming catastrophic forgetting in neural networks.
\newblock {\em Proceedings of the National Academy of Sciences (PNAS)},
  114(13):3521--3526, 2017.

\bibitem{li2018learning}
Z.~Li and D.~Hoiem.
\newblock Learning without forgetting.
\newblock {\em IEEE Transactions on Pattern Analysis and Machine Intelligence
  (PAMI)}, 40(12):2935--2947, 2018.

\bibitem{lin2014microsoft}
T.-Y. Lin, M.~Maire, S.~Belongie, J.~Hays, P.~Perona, D.~Ramanan,
  P.~Doll{\'a}r, and C.~L. Zitnick.
\newblock Microsoft coco: Common objects in context.
\newblock In {\em Proceedings of European Conference on Computer Vision
  (ECCV)}, pages 740--755. Springer, 2014.

\bibitem{lopez2017gradient}
D.~Lopez-Paz and M.~Ranzato.
\newblock Gradient episodic memory for continual learning.
\newblock In {\em Advances in Neural Information Processing Systems (NIPS)},
  pages 6467--6476, 2017.

\bibitem{mccloskey1989catastrophic}
M.~McCloskey and N.~J. Cohen.
\newblock Catastrophic interference in connectionist networks: The sequential
  learning problem.
\newblock In {\em Psychology of learning and motivation}, volume~24, pages
  109--165. Elsevier, 1989.

\bibitem{michieli2018game}
U.~Michieli and L.~Badia.
\newblock {Game Theoretic Analysis of Road User Safety Scenarios Involving
  Autonomous Vehicles}.
\newblock In {\em IEEE International Symposium on Personal, Indoor and Mobile
  Radio Communications}, pages 1377--1381, 2018.

\bibitem{nekrasov}
V.~Nekrasov.
\newblock Pre-computed weights for {ResNet-101}.
\newblock https://github.com/DrSleep/tensorflow-deeplab-resnet, Accessed:
  2019-07-04.

\bibitem{oquab2014learning}
M.~Oquab, L.~Bottou, I.~Laptev, and J.~Sivic.
\newblock Learning and transferring mid-level image representations using
  convolutional neural networks.
\newblock In {\em Proceedings of IEEE Conference on Computer Vision and Pattern
  Recognition (CVPR)}, pages 1717--1724, 2014.

\bibitem{polikar2001learn}
R.~Polikar, L.~Upda, S.~S. Upda, and V.~Honavar.
\newblock Learn++: An incremental learning algorithm for supervised neural
  networks.
\newblock {\em IEEE Transactions on Systems, Man, and Cybernetics, part C
  (Applications and Reviews)}, 31(4):497--508, 2001.

\bibitem{rebuffi2017icarl}
S.-A. Rebuffi, A.~Kolesnikov, G.~Sperl, and C.~H. Lampert.
\newblock icarl: Incremental classifier and representation learning.
\newblock In {\em Proceedings of IEEE Conference on Computer Vision and Pattern
  Recognition (CVPR)}, pages 2001--2010, 2017.

\bibitem{roy2018tree}
D.~Roy, P.~Panda, and K.~Roy.
\newblock {Tree-CNN: a hierarchical deep convolutional neural network for
  incremental learning}.
\newblock {\em arXiv preprint arXiv:1802.05800}, 2018.

\bibitem{sarwar2017incremental}
S.~S. Sarwar, A.~Ankit, and K.~Roy.
\newblock Incremental learning in deep convolutional neural networks using
  partial network sharing.
\newblock {\em arXiv preprint arXiv:1712.02719}, 2017.

\bibitem{shin2017continual}
H.~Shin, J.~K. Lee, J.~Kim, and J.~Kim.
\newblock Continual learning with deep generative replay.
\newblock In {\em Advances in Neural Information Processing Systems (NIPS)},
  pages 2990--2999, 2017.

\bibitem{shmelkov2017incremental}
K.~Shmelkov, C.~Schmid, and K.~Alahari.
\newblock Incremental learning of object detectors without catastrophic
  forgetting.
\newblock In {\em Proceedings of International Conference on Computer Vision
  (ICCV)}, pages 3400--3409, 2017.

\bibitem{dhar2018learning}
R.~V. Singh, P.~Dhar, K.-C. Peng, Z.~Wu, and R.~Chellappa.
\newblock Learning without memorizing.
\newblock {\em Proceedings of IEEE Conference on Computer Vision and Pattern
  Recognition (CVPR)}, 2019.

\bibitem{tasar2018incremental}
O.~Tasar, Y.~Tarabalka, and P.~Alliez.
\newblock {Incremental Learning for Semantic Segmentation of Large-Scale Remote
  Sensing Data}.
\newblock {\em arXiv preprint arXiv:1810.12448}, 2018.

\bibitem{thrun1996learning}
S.~Thrun.
\newblock Is learning the n-th thing any easier than learning the first?
\newblock In {\em Advances in Neural Information Processing Systems (NIPS)},
  pages 640--646, 1996.

\bibitem{wu2019}
Y.~Wu, Y.~Chen, L.~Wang, Y.~Ye, Z.~Liu, Y.~Guo, and Y.~Fu.
\newblock Large scale incremental learning.
\newblock In {\em Proceedings of IEEE Conference on Computer Vision and Pattern
  Recognition (CVPR)}, 2019.

\bibitem{wu2018incremental}
Y.~Wu, Y.~Chen, L.~Wang, Y.~Ye, Z.~Liu, Y.~Guo, Z.~Zhang, and Y.~Fu.
\newblock Incremental classifier learning with generative adversarial networks.
\newblock {\em arXiv preprint arXiv:1802.00853}, 2018.

\bibitem{xiao2014error}
T.~Xiao, J.~Zhang, K.~Yang, Y.~Peng, and Z.~Zhang.
\newblock Error-driven incremental learning in deep convolutional neural
  network for large-scale image classification.
\newblock In {\em Proceedings of the 22nd ACM International Conference on
  Multimedia (ACMMM)}, pages 177--186. ACM, 2014.

\bibitem{zhou2019M2KD}
P.~Zhou, L.~Mai, J.~Zhang, N.~Xu, Z.~Wu, and L.~S. Davis.
\newblock M2kd: Multi-model and multi-level knowledge distillation for
  incremental learning.
\newblock {\em Proceedings of IEEE Conference on Computer Vision and Pattern
  Recognition (CVPR)}, 2019.

\end{thebibliography}
}

\end{document}